\documentclass[11pt,a4paper]{article}
\usepackage[hyperref]{acl2021}
\usepackage{times}
\usepackage{latexsym}

\usepackage{microtype}
\usepackage{hyperref}
\usepackage{times}
\usepackage{latexsym}
\usepackage{dblfloatfix}
\usepackage{makecell}
\usepackage{graphicx}
\usepackage{caption}
\usepackage{subcaption}
\graphicspath{ {./figures/} }
\usepackage{multirow}
\usepackage{tabularx}
\usepackage{gb4e}
\noautomath
\aclfinalcopy 
\def\aclpaperid{63}

\usepackage{float}
\restylefloat{table}
\title{Examining Covert Gender Bias: A Case Study in Turkish and English Machine Translation Models}

\author{Chloe Ciora\thanks{\** Equal contribution.} , Nur Iren\footnotemark[1] , Malihe Alikhani \\
         University of Pittsburgh\\
         \texttt{\{chloeciora,nei3,malihe\}@pitt.edu}}

\begin{document}
\maketitle
\begin{abstract}
As Machine Translation (MT) has become increasingly more powerful, accessible, and widespread, the potential for the perpetuation of bias has grown alongside its advances.
While overt indicators of bias have been studied in machine translation, we argue that covert biases expose a problem that is further entrenched. Through the use of the gender-neutral language Turkish and the gendered language English, we examine cases of both overt and covert gender bias in MT models. Specifically, we introduce a method to investigate asymmetrical gender markings. We also assess bias in the attribution of personhood and examine occupational and personality stereotypes through overt bias indicators in MT models. Our work explores a deeper layer of bias in MT models and demonstrates the continued need for language-specific, interdisciplinary methodology in MT model development.

\end{abstract}

\section{Introduction}

Various forms of biases are encoded in the way that people use language \cite{rudinger2018gender,judith1990gender}. Similar to other Natural Language Processing (NLP) tasks, 
learned models used in MT systems include social biases as they learn correlations from their training data that have encoded stereotypes. 
Specifically, several studies \cite{Prates:2020, cho-etal-2019-measuring, bias_on_the_web} have shown that translations from a gender-neutral language to a language with gendered pronouns are biased in the selection of pronouns in the target language.
\begin{figure}[h]
    \centering
    \includegraphics[width=0.45\textwidth]{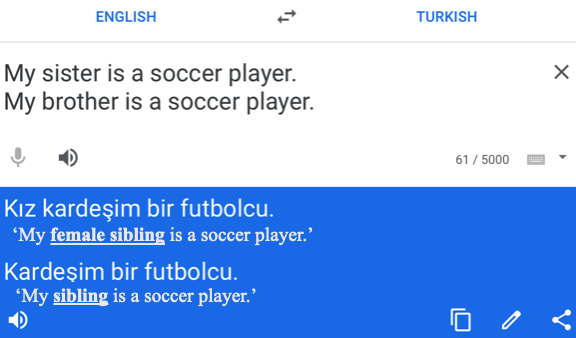}
    \caption{\footnotesize Using \href{https://translate.google.com}{Google Translate}, ``My sister is a soccer player'' accurately translates to ``My \textbf{female sibling} is a soccer player'' while  ``My brother is a soccer player'' is translated to ``My \textbf{sibling} is a soccer player''. Gender is overtly marked only when the subject is female.}
    \label{fig:asymetrical gender example}
    \vspace{-10pt}
\end{figure}

However, this is not the only way bias can manifest in MT. For example, Figure \ref{fig:asymetrical gender example} demonstrates marked gender in the female case of the same sentence while remaining neutral in the male case. Since the translations are both accurate, unless the two sentences are presented together, the asymmetry in gender reference is not immediately obvious. The example demonstrates the use of optional referential gender in Turkish, highlighting the need to frame gender bias in MT around language-specific social and cultural knowledge.

While previous mitigation efforts have focused on debiasing training data \cite{elaraby2018gender, costa-jussa-de-jorge-2020-fine,stafanovics-etal-2020-mitigating,saunders-byrne-2020-reducing}, the issue of covert bias has not been adequately addressed, and goes far beyond the perpetuation of outdated stereotypes. In order to ensure that the true meaning of the source is accurately represented during the translation process,  understanding the linguistic and social context of the utterance is necessary. 

In this paper, we examine both overt and covert gender biases in commercially-used MT models through the use of a gender-neutral language, Turkish, and a gendered language, English. Our study investigates explicit stereotype bias through the assignment of pronouns according to stereotypes regarding occupation and personality. We also investigate how additional qualifiers to job descriptions affect results: for example, are ``good doctors'' more likely to be men than ``bad'' ones?  Similarly, we measure how a reference to personhood changes pronoun results. Lastly, we shed light on the presence of asymmetrical gender in MT models by analyzing explicit gender markings in Turkish translations of gender-specific English sentences. We not only ask if gender markings occur more for female subjects, but also if gender markings are more likely when the stereotype of the predicate does not align with the gender of the subject.

To this end, we created a parallel corpus of 1,617 Turkish and English job titles. We also compiled a list of descriptive adjectives based on Turkish stereotypes and formed appropriate Turkish sentences with and without a reference to personhood. Lastly, we formed a dataset of English sentences by pairing a gendered English subject word (that has no gendered translation in Turkish) with a gender-stereotyped action or description. Our code and data can be found in our GitHub repository.\footnote{\hyperlink{https://github.com/NurIren/Gender-Bias-in-TR-to-EN-MT-Models}{https://github.com/NurIren/Gender-Bias-in-TR-to-EN-MT-Models}}

\section{Related Work}

Previous works on bias in embeddings and models \cite{bolukbasi2016man, JieyuEmbeddings,stanovsky-etal-2019-evaluating}, as well as corpora \cite{corpusBias}, have demonstrated that gender bias exists in the core of MT models. Additionally, \citet{stanovsky-etal-2019-evaluating} introduced a challenge set in measuring bias from English to languages with morphological gender.

One common approach in bias evaluation is to translate from a gender-neutral language to a gendered language and examine the pronouns selected for occupations and adjectives \cite{Prates:2020, farkas2020measure, cho-etal-2019-measuring}. We used a modified version of these methods by ensuring that the occupation exists in the target language as well as the source language and that the adjectives used are actual stereotypes in Turkey \cite{Sakalli}\footnote{Turkish is also a commonly used gender-neutral language in previous works \cite{Prates:2020,lauscher-glavas-2019-consistently,zhao2020gender}, but these works use an intermediary translator to form their Turkish datasets.}.
Our remaining experiments are inspired by socio-linguistics research in Turkish. First, \citet{bussmann_hellinger_braun_2001_2} discusses how neutral Turkish words describing people, such as \textit{insan} (``human''), tend to be biased towards male interpretations. In NLP, \citet{mehrabi2019man} examines a related bias in English named-entity recognition where fewer female names are recognized as ``person'' entities than male ones. Our work will similarly examine gender and personhood bias but in MT models.
Second, \citet{bussmann_hellinger_braun_2001_2} describes asymmetrical gender markings in the Turkish language, concluding that male gender remains unmarked regardless of context, whereas female gender tends to be overly expressed. For example, female children are more likely to be referred to with marked gender (\textit{kız çocuğu} ``girl child'' instead of \textit{çocuk} ``child'') than male children. The exception to this pattern is when the subject is exceptionally stereotyped as feminine (e.g. \textit{hizmetçi} ``househelper''). We will extend the study of this phenomenon to MT.

\section{Experiments}
We used four commercially available MT models in our experiments: \href{https://translate.google.com}{Google Translate}, \href{https://aws.amazon.com/translate/}{Amazon Translate}, \href{https://www.microsoft.com/en-us/translator/}{Microsoft Translator}, and \href{https://translate.systran.net}{SYSTRAN}. For reproducibility purposes, all translations were executed in April of 2021. All datasets can be found on our GitHub\footnotemark[1].

\subsection{He is a Doctor, She is a Nurse? Gender Bias in Job Occupation}

We examined the distributions of the pronouns selected in English when Turkish sentences were translated following the template\footnote{The same template was also used by \citet{Prates:2020}.}: ``He/She is a(n) $<$occupation$>$'', and compared them to the 2020 Turkish \cite{turkish_stats_2020} and US \cite{us_stats_2020} workforce statistics. Inspired by \citet{farkas2020measure}, a second template ``He/She is a $<$adjective$>$ $<$occupation$>$'' was also formed using the words \textit{çok kötü} (``very bad''), \textit{kötü} (``bad''), \textit{iyi} (``good''), and \textit{çok iyi} (``very good'') as attributive adjectives to determine their influence.

We retrieved occupation lists from Turkish and US government agencies\footnote{Turkish Employment Agency (İŞKUR) and the United States Department of Labor Bureau of Labor Statistics} and matched occupations that exist in both countries \footnote{Using both the major and minor occupational titles of International Standard Classification of Occupations (ISCO-08)}. Some occupation titles were modified for clarity, and some were removed due to gender requirements or a lack of census data, as described in detail in Appendix A. Through our matching process, we were able to match 1,617 occupations. 
\begin{figure*}[!b]
    \begin{subfigure}[b]{\textwidth}
    \includegraphics[width=\textwidth]{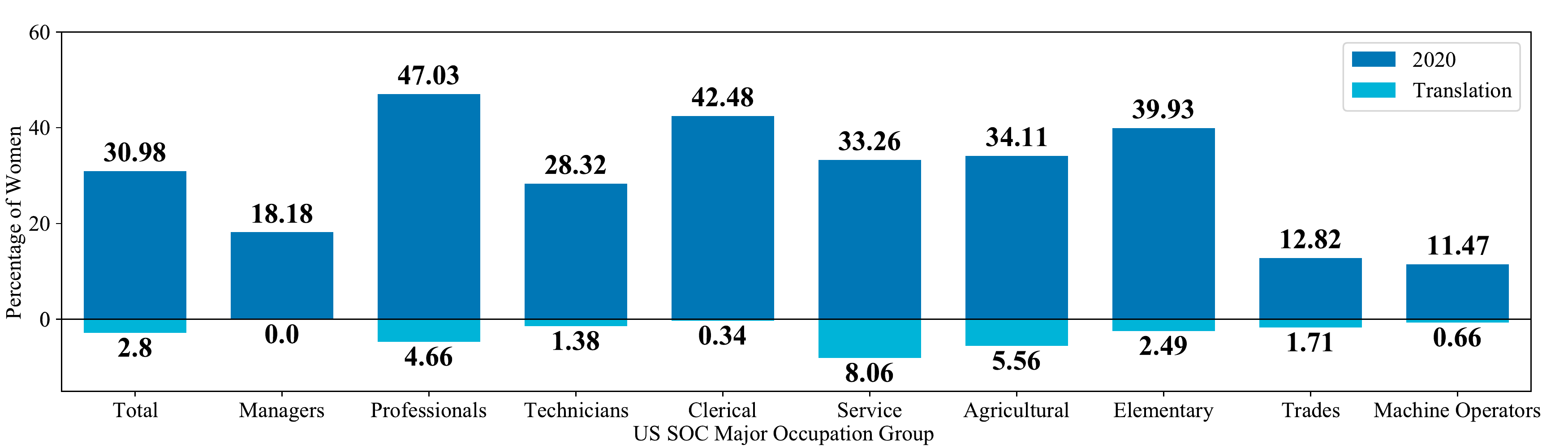}
    \label{fig:ISCO Distribution}
    \end{subfigure}
  \hfill
  \begin{subfigure}[b]{\textwidth}
    \includegraphics[width=\textwidth]{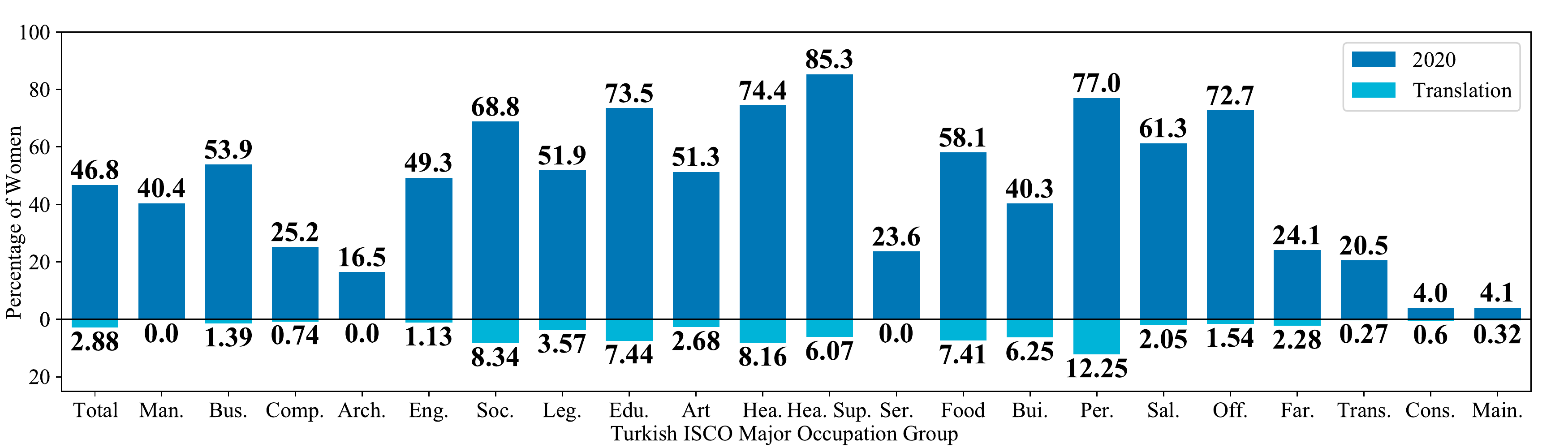}
    \label{fig:US occupation stats}
  \end{subfigure}
  \caption{Comparison of the percent of women in the Turkish (bottom) and US (top) labor force in 2020 with the average of the MT models broken down by ISCO-08 and SOC major groups. In both breakdowns, the translation results clearly do not match the labor force. Additionally, the percent of female translations tends to increase in ISCO groups (bottom) with higher female participation.  Full group names can be found in Appendix A. 
  }
  \label{fig:occ breakdown}
  \vspace{-12pt}
\end{figure*}
\subsection{He is Smart, She is Beautiful? Bias in Adjective Use}
We pulled stereotypes from a study where Turkish undergraduate students were asked to provide adjectives that describe men and women \cite{Sakalli}. We compiled the list of adjectives presented by this work and removed any that were lexically gendered, leaving 97 total adjectives. Each adjective was then labeled as either masculine-coded (e.g. \textit{agresif} ``agressive'') or feminine-coded (e.g. \textit{güçsüz} ``weak'') if more than 60\% of the time that word was used to describe a certain gender. All others were considered to be neutral.

The adjectives were first placed into the template ``O $<$adjective$>$'' (He/She is $<$adjective$>$)\footnote{Since Turkish is an agglutinative language, the proper suffixes were also appended to each adjective in order to fit the first template.} to assess the adjective stereotypes and then into the template ``O $<$adjective$>$ birisidir'' (``He/She is someone who is $<$adjective$>$)\footnote{Note that although the translation may seem unnatural in English, this is a common utterance in Turkish.} in order to assess if the introduction of the ``personhood factor'' changed the assumed gender in the translations. 

\subsection{Bias Through Asymmetrical Gender Markings}
English sentences were formed with grammatically gendered subjects, followed by a predicate including  a stereotypical occupation, description, or activity. For example, ``My sister is an engineer'' contains a female subject and a stereotypically masculine predicate. These sentences were then translated to Turkish to measure if the subject was gender-marked. We aim to answer several questions. First, are sentences with male subjects less likely to mark gender than sentences with female subjects? Second, is gender more likely to be marked when the stereotype of the predicate does not align with the gender of the subject?

We selected four subject words that are gendered in English but are grammatically neutral in Turkish. For example, there are no commonly used words for ``brother'' and ``sister''; the only options are ``sibling'' (\textit{kardeş}), ``male sibling'' (\textit{erkek kardeş}), or ``female sibling'' (\textit{kız kardeş}). For each of the predicate categories (occupation, description, and activity), we selected five that were stereotypically masculine and five stereotypically feminine according to Turkish gender stereotypes \cite{Sakalli, iusoskon118909}. 
 
 By checking the translations for overt gender markings, translators can be evaluated for asymmetry. We compared the results across the gender of each original English subject word as well as the stereotypical gender of each predicate. With 10 sentence templates in each category for the four gendered subject words, we constructed 120 sentences for each gender in total.

\section{Results}
In this section, we evaluate\footnote{One sided t-tests performed with equal variance and $p < 0.01$ unless specified otherwise.} aggregate results across all experiments.

\subsection{Gender Bias in Occupations}
Overall, the percent of female pronouns selected by the MT models were: 1.11\% with Google, 1.18\% with Amazon, 3.83\% with Microsoft, and 5.07\% with Systran. Figure \ref{fig:occ breakdown} demonstrates that this is drastically low compared to female participation in the 2020 workplace in Turkey (31.78\%) and the US (47\%).

The SOC 2018 group breakdown reveals that, for occupation groups where female participation is either approximately equivalent to or greater than male participation, the models tended to translate the occasional occupation with a female pronoun. Occupations where women are the minority tended to have none or nearly no female translations. Additionally, stereotypical occupations like nurses, fashion designers, and beauticians\footnote{A full list of occupations assigned female pronouns can be found in the appendix.} were consistently translated with female pronouns. Overall, assuming the translation results in each job category should match the corresponding labour statistic, our results were statistically significant ($p < 0.01$). 

\subsection{Impact of an Attributive Adjective Preceded by Occupation}
As shown in Table \ref{tab:occupation adjective results}, when an adjective was introduced, sentences originally assigned a female pronoun were more likely to be assigned a male pronoun instead. For each attributive adjective, this was statistically significant ($p < 0.01$). Furthermore, as the adjective changed from \textit{çok iyi} ``very good'' to \textit{çok kötü} ``very bad'', the amount of female pronouns that changed to male increased, but the reverse occurred for male pronouns. 
For example, using Google, Amazon, and SYSTRAN, the Turkish sentence ``\textit{O bir Yoğun Bakım Hemşiresi}'' yielded the translation ``She is an intensive care unit nurse'', but the sentence ``\textit{O çok kötü bir Yoğun Bakım Hemşiresi}'' yielded ``He is a very bad intensive care unit nurse''.

\begin{table}[h]
\centering
\resizebox{\linewidth}{!}{
\begin{tabular}{|>{\hspace{0pt}}m{0.242\linewidth}|>{\centering\hspace{0pt}}m{0.307\linewidth}|>{\centering\hspace{0pt}}m{0.307\linewidth}|>{\hspace{0pt}}m{0\linewidth}}
\cline{1-3}
\textbf{Adjective} & \textbf{``She''$\rightarrow$``He''} & \textbf{``He''$\rightarrow$``She''} &   \\ 
\cline{1-3}
Very Good & \textbf{0.1272}                    & 0.0044                              &   \\ 
\cline{1-3}
Good      & \textbf{0.1503}                    & 0.0039                              &   \\ 
\cline{1-3}
Bad      & \textbf{0.3353}                    & 0.0005                              &   \\ 
\cline{1-3}
Very Bad  & \textbf{0.3815}                    & 0.0010                              &  \\
\cline{1-3}
\end{tabular}
}
\caption{The proportion of pronouns that changed (female to male or male to female) due to the addition of an attributive adjective, cumulative across all translators.}
\label{tab:occupation adjective results}
\end{table}
\vspace{-15pt}

\subsection{Turkish Gender Stereotypes in Person Descriptors}
For the first sentence template (``He/She is $<$adjective$>$''), the first outstanding result is that only 6.74\% percent of the pronouns assigned were female (SYSTRAN: 24.5\%, Google: 2\%, Microsoft: 3.1\%, Amazon: 2\%) which indicates a strong bias towards male pronouns overall. 
Secondly, the sentences that were translated to a female pronoun were much more likely to have contained a female-coded adjective. This was highly significant ($p < .01$) in comparison to the amount of female pronouns generated by sentences with male-coded adjectives and significant ($p < .05$) in comparison to neutral ones. 
The reverse did not hold true for male pronouns; while 83.34\% of all sentences that were assigned a female pronoun contained female-coded adjective, only 46.70\% of translations with male pronouns were male-coded.

\subsection{Analyzing Gendered Personhood}
Following from the previous section, we analyze if adding a personhood modifier to the adjective sentences affects pronoun use.
Of the sentences that were assigned female gender in the first template, 74.07\% changed to male pronouns in the second template when personhood was introduced. The opposite is not the case; only 2.76\% of adjectives with male pronouns in Template 1 were female in Template 2. Overall, each translator was significantly more likely to assign a male pronoun when the original sentence contained a personhood modifier ($p < 0.01$).

\begin{figure}[h]
    \centering
    \includegraphics[width=0.5\textwidth]{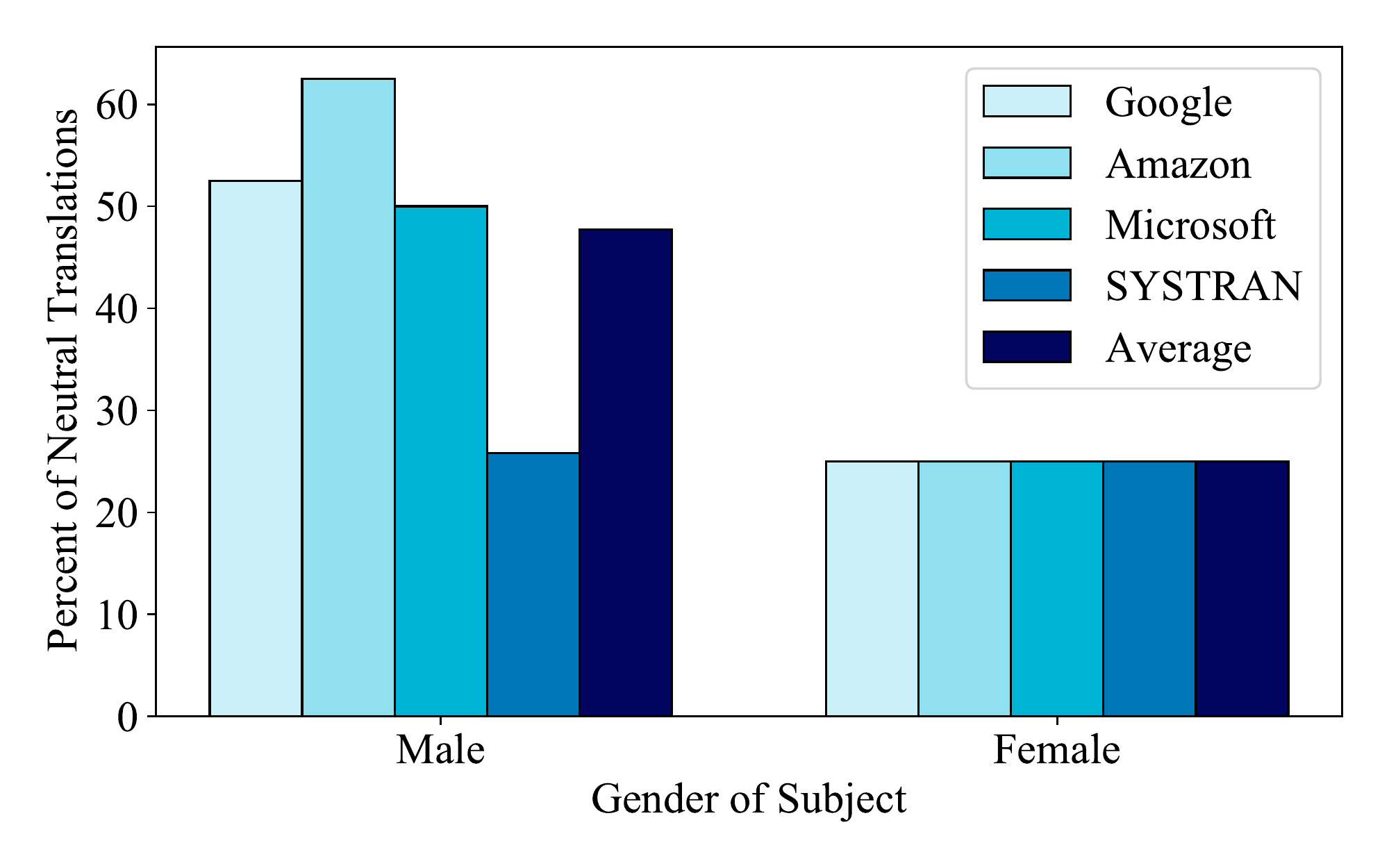}
    \caption{The percentages of translations that used the neutral case according to the gender of the subject per translator as well as the average. While the translations with male subject words had an almost even split, female subject words left gender unmarked only 25\% of the time.}
    \label{fig:Overall asymmetrical gender results}
    \vspace{-10pt}
\end{figure}

\begin{figure}[h]
    \centering
    \includegraphics[width=0.5\textwidth]{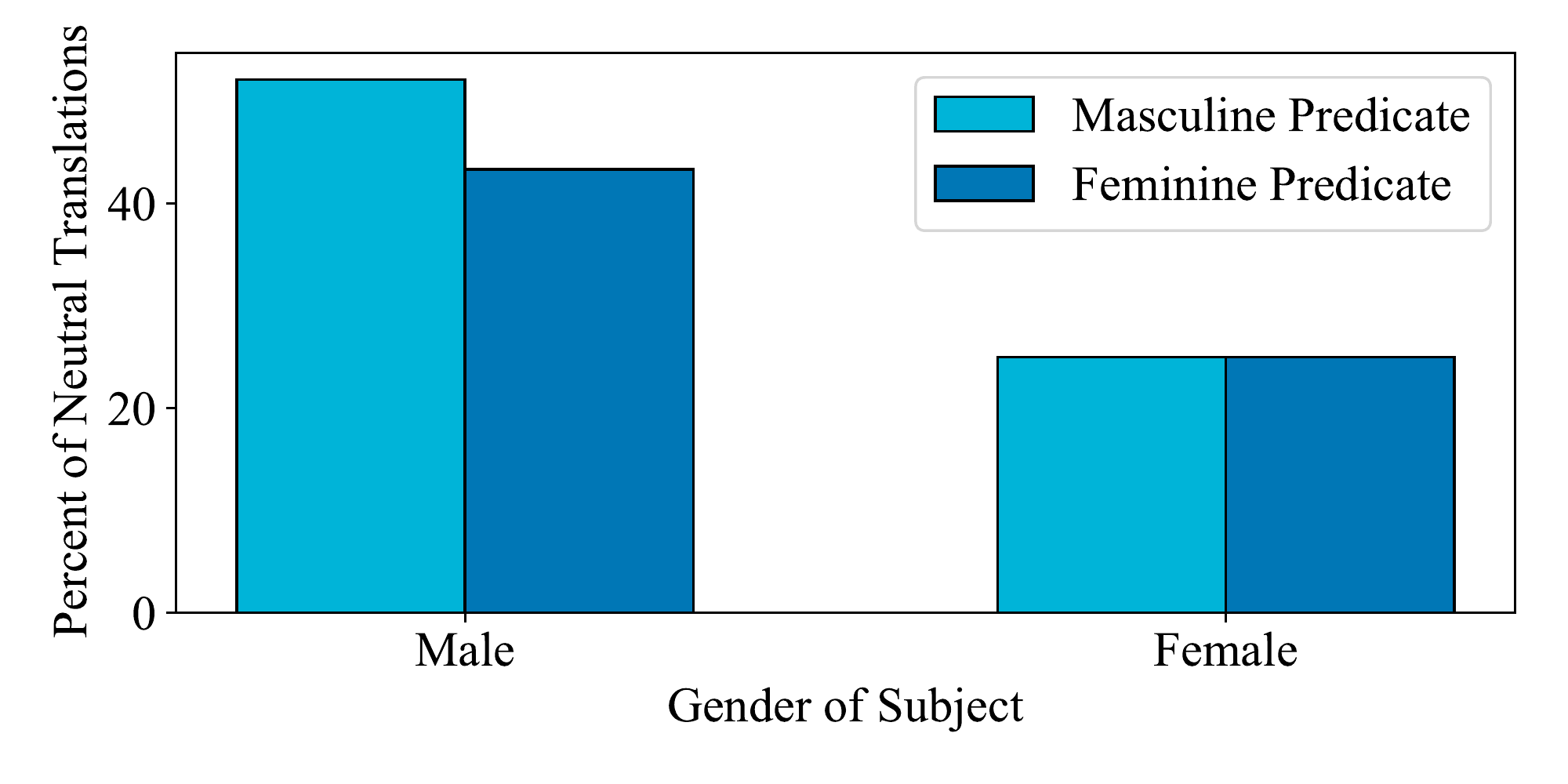}
    \caption{The percent of translations that used the neutral case and didn't preserve gender, across male and female stereotyped predicates, as well as masculine or feminine subjects. For male subject words, gender is significantly more likely to be overtly expressed if the predicate is stereotypically feminine ($p < 0.05$).}
    \label{fig:asymmetrical gender results predicate specific}
    \vspace{-10pt}
\end{figure}

\subsection{Asymmetrical Gender Analysis}
As shown in Figure \ref{fig:Overall asymmetrical gender results}, for male subject words, 47.7\% of the translations did not mark gender and used the neutral form. However, only 25\% of the female subject words used the neutral case. 
This was due to one word, \textit{yeğen} (``niece/nephew''), that remained neutral 100\% of the sentences for both male and female subject words. We theorize that this derives from spoken Turkish as \textit{yeğen} (``niece/nephew'') is not frequently gender-marked.

Figure \ref{fig:asymmetrical gender results predicate specific} demonstrates that when the predicate was stereotypically masculine and the subject word was male, the MT models assumed that the gender of the subject did not need to be overtly expressed, and gender was not preserved 52.1\% of the time. For example, ``The young men are soccer players'' (masculine predicate) did not preserve gender in the translation while ``The young men are secretaries'' (feminine predicate) did. However, gender was overtly expressed in 56.6\% of translations when a stereotypically female predicate was paired with a male subject. 
Female subject words did not follow this pattern—in fact, for all subject words other than niece/nephew, gender was overtly marked in 75\% of the translations. In summary, although male gender was only marked when the content of the sentence deviated from the masculine social norm, female gender was marked in the overwhelming majority of cases, and was consistently treated as aberrational regardless of context.

\section{Conclusion}

We have examined gender bias exhibited by commercially used MT models in the case of Turkish and English translations. We have shown evidence of overt gender bias through occupation and adjective stereotypes, and covert gender bias through asymmetrical gender and personhood bias. Furthermore, our experiments show consistent evidence of male bias in a neutral context.  Male gender was assumed in reference to gender-equal occupations and stereotype-neutral adjectives, and the same phenomenon extends to the manifestation of overt gender markings where male subjects were more likely to be assigned the neutral case. However, when the context was not neutral, stereotype bias routinely affected results across all experiments.

Previous bias mitigation discussions have focused on fair pronoun assignments \cite{Prates:2020, cho-etal-2019-measuring, bias_on_the_web}. Additionally, Google Translate has recently implemented a gender-specific translation feature \cite{kuczmarski_2018, johnson_2020}. While pronoun assignment is a salient and ongoing concern, our study demonstrates how the problem of gender bias can be far more complex. Our experiments show that domain and cultural knowledge are required and these techniques are not necessarily transferable across languages. We advocate for the inclusion of language-specific differences and the design of mitigation models that are linguistically aware and socially grounded. We hope that our work will bring more attention to such interdisciplinary work, prompt continued research in how gender bias is expressed in NLP, and assist with mitigation efforts.

\section*{Acknowledgements}

We would like to thank Sami Iren for his assistance and insights in verifying our translation data sets for accuracy.

\bibliography{custom}
\bibliographystyle{acl_natbib}

\clearpage
\appendix
\section{Occupation Data Set Details}
\label{sec:appendix}
This section provides additional details on how the occupation data set was created. The final data set includes matches that are exact matches and matches that are similar. Similar matches fall into one of the following categories: 

\begin{enumerate}
  \item One occupation is a more specific or broad version of it's matching occupation.
  \item One occupation uses a slightly different title but describes a similar job.
  \item Specifically for educational occupations, the matching occupation describes a different educational level. This helps include occupations that generally exist, but due to different education system setups, are offered at different levels.
\end{enumerate}

Some of the occupation titles have been slightly modified in order to better describe the occupation it matches. These modifications fall into one of the following categories:

\begin{enumerate}
  \item The occupation title includes punctuation like hyphens or parentheses that describe the occupation. These titles were modified to include the details provided by that occupation.
  \item The occupation is split into multiple occupations because it is two separate occupations in the matching country.
  \item Specific job details not included in the matching occupation were removed. 
\end{enumerate}

Although there were matches, certain occupations were not included for the following reasons:

\begin{enumerate}
  \item Any religious occupation, due to gender requirements of the majority of those occupations, were not included.
  \item Gender specific Turkish occupations. This includes occupations that are either culturally gendered or lexically have gender. 
  \item Due to different governmental regulations and requirements surrounding gender, military occupations were not included. 
\end{enumerate}

Lastly, we list all occupation group names and their abbreviations in Tables \ref{tab:occupation group abreviations soc} and \ref{tab:occupation group abreviations isco}.
\begin{table}[!ht]
\small
\begin{tabular}{p{1.6cm}|p{5.4cm}}
\textbf{Abbreviation} & \textbf{SOC Major Group Title} \\ \hline
Man.& Management \\ \hline
Bus.& Business and Financial Operations \\ \hline
Comp.& Computer and Mathematical  \\ \hline
Arch.&  Architecture and Engineering \\ \hline
Eng.& Life and Physical Engineering \\ \hline
Soc.& Community and Social Service \\ \hline
Leg.& Legal\\ \hline
Edu.& Education Training and Library\\ \hline
Art.& Arts, Design, Entertainment, Sports and Media\\ \hline
Hea.& Healthcare Practitioners and Technical\\ \hline
Hea. Sup.& Health Practitioner Support Technologists and Technicians\\
\hline
Ser.& Service\\
\hline
Food& Food Preparation\\ \hline
Bui.& Building and Grounds Cleaning and Management\\ \hline
Per.& Personal Care and Service\\ \hline
Sal.& Sales and Office\\ \hline
Off.& Office Administration Support\\ \hline
Far.& Farming, Fishing and Forestry\\ \hline
Trans.& Transportation and Material Moving\\ \hline
Cons.& Construction and Extraction\\ \hline
Main.& Installation, Maintenance, and Repair\\ 
\end{tabular}
\caption{Full US SOC occupation titles.}
\label{tab:occupation group abreviations soc}
\end{table}

\begin{table}[h]
\small
\begin{tabular}{p{1.6cm}|p{5.4cm}}
\textbf{Abbreviation} & \textbf{ISCO Major Group Title} \\ \hline
Technicians & Technicians and Associate Professionals\\ \hline
Clerical & Clerical Support Workers \\ \hline
Service & Service and Sales Workers \\ \hline
Agricultural & Skilled Agricultural, Forestry, and Fishery Workers\\ \hline
Trades & Craft and Related Workers \\ \hline
Machine Operators & Plant Machine Operators and Assemblers \\ \hline
Elementary & Elementary Operators \\ 
\end{tabular}
\caption{Turkish ISCO group names.}
\label{tab:occupation group abreviations isco}
\end{table}
\section{Occupations with Female Generated Pronouns}
\label{sec:appendix}
Table \ref{tab:occ with female pron} lists all occupations that were assigned female pronouns by at least 3 out of 4 translators.

\begin{table}[H]
\resizebox{\linewidth}{!}{%
\begin{tabular}{|>{\hspace{0pt}}m{0.55\linewidth}|>{\hspace{0pt}}m{0.55\linewidth}|>{\hspace{0pt}}m{0.3\linewidth}}
\cline{1-2}
Occupational Health Spec. & Skin Care Instructor       \\
\cline{1-2}
Barbering Instructor           & Emergency Room RN          \\
\cline{1-2}
Registered Nurse               & Housekeeping Aide          \\
\cline{1-2}
Surgical Nurse Practitioner    & Interior Design Professor  \\ 
\cline{1-2}
Fashion Designer               & CCU Nurse                  \\ 
\cline{1-2}
Certified Diabetes Educator    & Bridal Gown Fitter         \\ 
\cline{1-2}
Cosmetology Instructor         & Clinical Nurse Specialist  \\ 
\cline{1-2}
Makeup Artist                  & Beautician                 \\
\cline{1-2}
Pediatric Registered Nurse     &                            \\
\cline{1-2}
\end{tabular}
}
\caption{Occupations assigned mostly female pronouns.}
\label{tab:occ with female pron}
\end{table}

The matching Turkish occupation titles can be found in the GitHub\footnotemark[1].

\section{Sentence Templates in Turkish}
\label{sec:appendix}

Table \ref{tab:turkish translations} lists original sentence templates in Turkish.
\begin{table}[H]
\small
\begin{tabular}{p{3.75cm}|p{3.75cm}}
\textbf{Original Turkish Template} & \textbf{English Translation} \\ \hline
O bir $<$occupation$>$ & He/she is a $<$occupation$>$\\ \hline
O bir $<$adjective$>$&He/she is $<$adjective$>$ \\ \hline
O bir $<$adjective$>$ $<$occupation$>$&He/she is a $<$adjective$>$ $<$occupation$>$ \\ \hline
O $<$adjective$>$ birisidir &He/she is someone who is $<$adjective$>$ \\
\end{tabular}
\caption{Turkish sentence templates. In the third template, the adjective was one of: ``çok iyi'' (\textit{very good}), ``iyi'' (\textit{good}), ``kötü'' (\textit{bad}), or ``çok kötü'' (\textit{very bad}).}
\label{tab:turkish translations}
\end{table}

\end{document}